\documentclass[twoside,leqno,twocolumn]{article}

\usepackage[letterpaper]{geometry}

\usepackage{ltexpprt}

\usepackage{wrapfig}
\usepackage{booktabs} 
\usepackage{soul}
\usepackage{paralist}
\usepackage{multirow}
\usepackage{tabularx}
\usepackage{balance}
\usepackage{enumitem}
\usepackage{graphicx} 
\usepackage{subfigure}
\usepackage[skip=7pt]{caption}
\usepackage{xcolor}

\newtheorem{problem}{Problem}

\usepackage{bm}
\usepackage{amsmath}
\usepackage{amsfonts}
\usepackage{dsfont}
\usepackage{ amssymb }

\usepackage{algorithm}

\usepackage{algorithmic}

\usepackage{tikz-qtree}
\usepackage{tikz-qtree-compat}
\newcommand\independent{\protect\mathpalette{\protect\independenT}{\perp}}
\def\independenT#1#2{\mathrel{\rlap{$#1#2$}\mkern2mu{#1#2}}}

\usetikzlibrary{shapes,decorations,arrows,calc,arrows.meta,fit,positioning}
\tikzset{
	-Latex,auto,node distance =1 cm and 1 cm,semithick,
	state/.style ={ellipse, draw, minimum width = 0.7 cm},
	point/.style = {circle, draw, inner sep=0.04cm,fill,node contents={}},
	bidirected/.style={Latex-Latex,dashed},
	el/.style = {inner sep=2pt, align=left, sloped}
}

\begin{document}
\title{Counterfactual Evaluation of Treatment Assignment Functions with Networked Observational Data}
\author{Ruocheng Guo\thanks{Arizona State University, Computer Science and Engineering} \and Jundong Li \thanks{University of Virginia, Department of Electrical and Computer Engineering, Department of Computer Science, and School of Data Science}
\and Huan Liu\footnotemark[1]
}
\date{}


\fancyfoot[R]{\footnotesize{\textbf{Copyright \textcopyright\ 20XX by SIAM\\
Unauthorized reproduction of this article is prohibited}}}



\maketitle

\begin{abstract}
Counterfactual evaluation of novel treatment assignment functions (e.g., advertising algorithms and recommender systems) is one of the most crucial causal inference problems for practitioners.
Traditionally, randomized controlled trials (A/B tests) are performed to evaluate treatment assignment functions.
However, such trials can be time-consuming, expensive, and even unethical in some cases.
Therefore, offline counterfactual evaluation of treatment assignment functions becomes a pressing issue because a massive amount of observational data is available in today's big data era.
Counterfactual evaluation requires handling the hidden confounders -- the unmeasured features which causally influence both the treatment assignment and the outcome.
To deal with the hidden confounders, most of the existing methods rely on the assumption of no hidden confounders.
However, this assumption can be untenable in the context of massive observational data.
When such data comes with network information, the later can be potentially useful to correct hidden confounding bias.
%
As such, we first formulate a novel problem, counterfactual evaluation of treatment assignment functions with networked observational data.
Then, we investigate the following research questions: How can we utilize network information in counterfactual evaluation? Can network information improve the estimates in counterfactual evaluation?
	Toward answering these questions, first, we propose a novel framework, \emph{Counterfactual Network Evaluator} (CONE), which (1) learns partial representations of latent confounders under the supervision of observed treatments and outcomes; and (2) combines them for counterfactual evaluation.
	Then through extensive experiments, we corroborate the effectiveness of CONE. The results imply that incorporating network information mitigates hidden confounding bias in counterfactual evaluation.
	
\end{abstract}

\section{Introduction}
The advanced big data technologies have granted us the convenient access to massive observational data in a plethora of highly influential applications such as social networks~\cite{guo2019learning}, online advertising, and recommender systems~\cite{schnabel2016recommendations}.
For example, an online blogger community generates massive log data containing bloggers' keywords (features), users' browsing devices (the treatments), and users' opinions (the outcomes).
With such data, a fundamental problem of causal inference is counterfactual evaluation of treatment assignment functions which assigns a treatment to each instance based on its features.
%
%
The goal of counterfactual evaluation is to estimate the utility of a treatment assignment function, i.e., the average outcome over a population under the treatments assigned by this function, without performing randomized controlled trials (RCTs).
This is because RCTs could be costly, time consuming, and even unethical to perform~\cite{pearl2009causal,guo2018survey}.
%
For example, an online blogging service company wants to estimate the utility of a new advertising algorithm that determines which browsing device to promote a blogger's articles based on her keywords.
To evaluate the algorithm, an A/B test, the de-facto of RCTs in this context, often takes at least two weeks to provide statistically significant results~\cite{ju2019sequential}. 
Therefore, alternative approaches are sought by data scientists who wish to leverage the massive log data from existing advertising algorithms to estimate the utility of a new algorithm.
%
	
%
	%
%

There are two important components that often come with massive observational data: (1) the hidden confounders -- the unmeasured features that causally influences treatment assignments and outcomes at the same time (e.g., the writing style of a blogger which influences readers' choice of browsing devices and their opinions); and (2) auxiliary network information (e.g., a social network of bloggers).
%
Counterfactual evaluation of treatment assignment functions with observational data often requires handling the influence of hidden confounders.
%
%
This is because estimates of a treatment assignment function's utility can be biased without controlling the influence of hidden confounders, even if the effect of observed features have been adjusted for.
A vast majority of existing methods~\cite{dudik2011doubly,athey2017efficient,qian2011performance,kallus2018balanced,zou2019focused} rely on the assumption of no hidden confounders (a.k.a. \emph{unconfoundedness}).
However, in the context of massive observational data this assumption can be untenable because it is unlikely to have all confounders measured.
Bennett and Kallus~\cite{bennett2019policy} proposed to relax the unconfoundedness assumption with proxies of the latent confounders.
However, the existing counterfactual evaluation methods cannot utilize the rich network information among observation data.

The importance of network structures in catching hidden confounders' patterns have been shown in other tasks of causal inference~\cite{veitch2019using,guo2019learning}.
For example, while the writing style of a blogger is hard to quantified, her representation can be implicitly captured by her social network patterns such as which bloggers are her friends.
However, how to facilitate counterfactual evaluation with network information remains untouched.
%
%
%
%
%

To bridge the gap, we propose to use network structural patterns to mitigate the hidden confounding bias in the task of counterfactual evaluation.
Specifically, we propose a novel framework -- \textbf{CO}unterfactual \textbf{N}etwork \textbf{E}valuator (CONE), which recognizes the patterns of hidden confounders from network structures for counterfactual evaluation with massive observational data.
The main contributions of this work are:
\begin{itemize}
    \item Formulating a novel research problem, counterfactual evaluation of treatment assignment functions with networked observational data;
    \item Proposing a framework, CONE, that provides a principled way to exploit the network information and the observed features for counterfactual evaluation with networked observational data; and
    \item Conducting extensive experiments to corroborate the effectiveness of CONE for counterfactual evaluation with networked observational data.
\end{itemize}
%

%

\section{Problem Statement}

\label{sec:problem}

In this section, we introduce the technical preliminaries and present the problem statement.
%

First, we start with the notations. Lower alphabets (e.g., $y_i$) denote scalars, uppercase alphabets (e.g., $N$) signify constants, lower and upper boldface alphabets (e.g., $\bm{x}$ and $\bm{A}$) denote vectors and matrices.
%
%

%
In networked observational data, each instance $i$ comes with a feature vector $\bm{x}_i \in \mathds{R}^M$, an observed treatment $t_i\in\{0,1\}$ and an observed (factual) outcome $y_i\in \mathds{R}$.
Besides, we observe a network connecting the instances, represented by its adjacency matrix $\bm{A}\in\{0,1\}^{N\times N}$.
To define the utility of a treatment function, we assume that there exists a potential outcome corresponding to each treatment-instance pair $(t,i)$, i.e., $y_i(1)$ and $y_i(0)$~\cite{rubin2005causal}.
For each instance $i$, the observed outcome $y_i$ (short for $y_i(t_i)$) takes the value on one of the potential outcomes, depending on the observed treatment ($t_i$).
Formally, this can be written as $y_i(t_i) = t_iy_i(1)+(1-t_i)y_i(0)$.
%
%
Following the literature~\cite{pearl2009causal}, we call the unobserved outcomes $y_i(t),\;t\not=t_i$ the \textit{counterfactual outcomes}.

%

%
%
%

%
As in almost all the cases, it is not possible to test whether the set of observed features contain all the confounders.
Therefore, we adopt a realistic setting where hidden confounders exist.
Thus, the unconfoundedness assumption does not hold given observed features:
\begin{equation}
    y_i(1), y_i(0) \not\independent t_i | \bm{z}_i.
\end{equation}
Instead, we only assume that there exist latent confounders $\bm{z}$ which satisfy
\begin{equation}
        y_i(1), y_i(0) \independent t_i | \bm{z}_i.
\end{equation}
Note that the latent confounders are not observable in the data.
But we can approximate them from the observed features and the network information.

Similar to its counterpart for i.i.d. data~\cite{bennett2019policy,athey2017efficient}, in networked observational data, a treatment assignment function $\pi:\mathds{R}^M\times \mathcal{A} \rightarrow (0,1)$ maps an instance's feature vector to its probability to receive the treatment, where $\mathcal{A}$ is the set of possible adjacency matrices.
Then $\pi^t(\bm{x})$ denotes the probability that treatment $t$ assigned to an instance with features $\bm{x}$ by $\pi$.
In accordance with~\cite{kallus2018balanced,bennett2019policy,zou2019focused}, we define the true utility of a treatment assignment function $\pi$ as:
\begin{equation}
\label{eq:utility}
\tau(\pi) = \frac{1}{N}\sum_i\sum_t\pi^t(\bm{x}_i,\bm{A})y_i(t).
\end{equation}
From this definition, we make the observation: when $t\not=t_i$, $y_i(t)$ is a counterfactual outcome which is not available in observational data. 
%
Therefore, counterfactual evaluation is a challenging problem.
Here, we formulate the problem studied in this work.
%
%
%
%
%
%
%
%
Based on the aforementioned notations and definitions, a formal statement of the problem is given as follows:

\begin{problem}{Counterfactual Evaluation of the Treatment Assignment Functions with Networked Observational Data.}
\begin{description}
\item[Given:] networked observational data $(\left\{(\bm{x}_i,{t}_i,y_i)\right\}_{i=1}^N,\bm{A})$ and a novel treatment assignment function $\pi$.
\item[Estimate:] its true utility $\tau(\pi)$ on the given data.
\end{description}
\end{problem}

\section{Background}

%
Here, we review three types of long-established approaches for counterfactual evaluation with independent and identically distributed (i.i.d.) observational data: direct methods~\cite{beygelzimer2008importance}, weighted estimators~\cite{bottou2013counterfactual,swaminathan2015counterfactual}, and doubly robust estimators~\cite{dudik2011doubly}.
In this section, the symbol $\pi$ is used to denote a treatment assignment function for i.i.d. observational data as $\pi:\mathds{R}^M \rightarrow \mathds{R}$.
Given i.i.d. observational data $\left\{\bm{x}_i,t_i,y_i\right\}_{i=1}^N$ and policy $\pi$, the directed method~\cite{qian2011performance} estimates the policy value as:
\begin{equation}
	\hat{\tau}(\pi) = \frac{1}{N}\sum_{i=1}^N\sum_t \pi^t(\bm{x}_i)\hat{y}_i(t),
	\label{eq:dm}
\end{equation}
where $\hat{y}_i(t)$ is the estimated outcome of instance $i$ under treatment $t$.
However, relying on the unconfoundedness assumption makes direct methods suffer from the bias caused by hidden confounders~\cite{beygelzimer2008importance}. 

Alternatively, the weighted estimators~\cite{bottou2013counterfactual,swaminathan2015counterfactual} are proposed to achieve counterfactual evaluation:
\begin{equation}
	\hat{\tau}(\pi) = \frac{1}{N}\sum_{i=1}^N \hat{w}(\bm{x}_i,t_i)y_i,
\end{equation}
where $\hat{w}(\bm{x}_i,t_i)$ maps the observed features and treatment of an instance to its weight.
In weighted estimators, the utility of a treatment assignment function is estimated by a weighted average of factual outcomes. 
As a result, weighted estimators do not need to bother with counterfactual outcomes.
Weighted estimators often adopt the inverse propensity scoring (IPS) weights~\cite{kitagawa2018should}:
\begin{equation}
	\hat{w}_{IPS}(\bm{x}_i,t_i) = \frac{\pi^{t_i}(\bm{x}_i)}{P(t=t_i|\bm{x})},
\end{equation} 
where $P(t=t_i|\bm{x}_i)$ denotes the true probability of instance $i$ to receive treatment $t_i$ in the observational data.
However, we often have to estimate $P(t=t_i|\bm{x}_i)$ as how the treatments are assigned in the observational data is unknown.
%
%
%
To avoid the extreme values of $\pi^t(\bm{x}_i)$, techniques including normalization and clipping have been introduced~\cite{bottou2013counterfactual,swaminathan2015counterfactual,swaminathan2015self}.

Moreover, the doubly robust estimator~\cite{dudik2011doubly} estimates utility of treatment assignment functions via utilizing both estimated counterfactual outcomes and IPS weights. It is defined as:
\begin{equation}
\hat{\tau}(\pi) = \frac{1}{N}\sum_{i=1}^N[ \sum_t \pi^t(\bm{x}_i)\hat{y}_i(t) + \hat{w}_{IPS}(\bm{x}_i,t_i)(y_i-\hat{y}_i(t_i))].
\label{eq:dr}
\end{equation}
%
%
We can see the three types of counterfactual evaluation methods cannot utilize the network information.
The success of using network information to handle hidden confounders has been demonstrated in other tasks of causal inference~\cite{veitch2019using,guo2019learning} (e.g., causal effect estimation). Motivated by the success, we investigate incorporating network information in counterfactual evaluation.
%


\section{Counterfactual Network Evaluator (CONE)}
\begin{figure}
\centering
\includegraphics[width=0.47\textwidth]{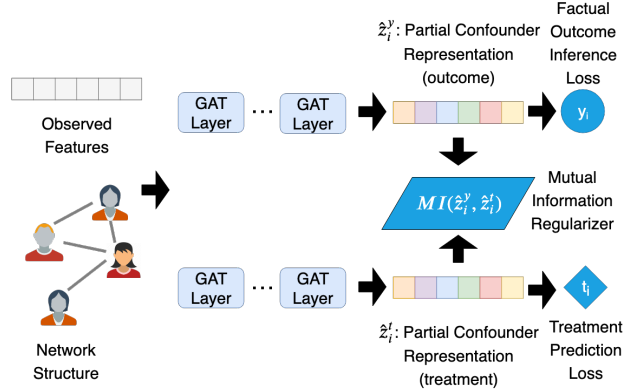}
\caption{An overview of learning partial representations of latent confounders in CONE.} \label{fig:model}
\end{figure}

In this section, we present the proposed framework to tackle the counterfactual evaluation problem with networked observational data.
 shows an overview of the proposed framework's training phase.
As shown in Fig.~\ref{fig:model}, the goal of the training phase is to learn two partial representations of latent confounders with the supervision of the factual outcome and the observed treatment, respectively.
Then in the inference phase, the learned partial representations would be utilized to infer the utility of a treatment assignment function.
Then we cover the detailed description of the proposed framework in the rest of this section.

\subsection{Learning Partial Repretations of Latent Confounders.}
To leverage the network information in the procedure of learning latent confounders' partial representations, for each partial representation, a representation learning function $g:\mathds{R}^M \times \mathcal{A} \rightarrow \mathds{R}^D$ maps the observed features along with the network information to the $D$-dimensional space of partial latent confounders.
We let $g^t$ and $g^y$ denote the partial representation learning functions supervised by the observed treatment and the factual outcome, respectively.
In this work, we approximate the functions, $g^t$ and $g^y$, with Graph Attentional (GAT) layers~\cite{velickovic2018graph} to capture the unknown edge weights in the real-world networked observational data.
Intuitively, each GAT layer maps a feature vector and the network information to a partial representation vector.
In this work, each GAT layer employs multi-head graph attention.
To compute a partial representation vector of instance $i$, it concatenates the multiple heads' outputs.
Each head outputs a weighted aggregation of information from the neighbors of instance $i$ in the network $\bm{A}$~\cite{velickovic2018graph}.
An arbitrary number of GAT layers can be stacked to approximate the functions $g^y$ and $g^t$. Here, for notation simplicity, each partial representation learning function is formulated by a single GAT layer as:
\begin{equation}
\begin{split}
        \hat{\bm{z}^t_i} = g^t(\bm{x}_i,\bm{A}) = \parallel_{k=1}^K \delta(\sum_{j\in\mathcal{N}_i}\alpha_{ij}^k\bm{W}^k\bm{x}_j) \\
        \hat{\bm{z}^y_i} = g^y(\bm{x}_i,\bm{A}) = \parallel_{k=1}^K \delta(\sum_{j\in\mathcal{N}_i}\beta_{ij}^k\bm{U}^k\bm{x}_j),
\end{split}
    \label{eq:pr}
\end{equation}
where $\parallel$ denotes concatenation.
$\mathcal{N}_i$ signifies the set of neighbors of the $i$-th instance in the network $\bm{A}$.
$K$ is the number of attention heads.
$\bm{W}^k,\bm{U}^k$ are the weight matrices of the $k$-th attention head.
$\delta$ is the ELU activation unit.
$\alpha_{ij}^k$ and $\beta_{ij}^k$ are the normalized attention coefficients which represent the importance of the edge between instance $i$ and $j$ in the inference of the observed treatment and outcome, respectively.
We compute them as:
\begin{equation}
\begin{split}
    \alpha_{ij}^k = \frac{\exp(\delta'(\bm{a}^T[\bm{W}^k\bm{x}_i\parallel\bm{W}^k\bm{x}_j]))}{\sum_{l\in\mathcal{N}_i}\exp(\delta'(\bm{a}^T[\bm{W}^k\bm{x}_i\parallel\bm{W}^k\bm{x}_l]))}, \\
        \beta_{ij}^k = \frac{\exp(\delta'(\bm{b}^T[\bm{U}^k\bm{x}_i\parallel\bm{U}^k\bm{x}_j]))}{\sum_{l\in\mathcal{N}_i}\exp(\delta'(\bm{b}^T[\bm{U}^k\bm{x}_i\parallel\bm{U}^k\bm{x}_l]))},
\end{split}
\end{equation}
where $\delta'$ denotes the LeakyReLU unit~\cite{xu2015empirical} and $\bm{a},\bm{b}\in\mathds{R}^{2M}$ denotes the weight vectors.
One can interpret $\delta'(\bm{a}^T[\bm{W}^k\bm{x}_i\parallel\bm{W}^k\bm{x}_j])$ and $\delta'(\bm{b}^T[\bm{U}^k\bm{x}_i\parallel\bm{U}^k\bm{x}_j])$ as the unnormalized attention coefficients of the edge between the instances $i$ and $j$. 
Then these coefficients are normalized by applying the softmax function.

\noindent\textbf{Factual Outcome Inference Loss.}
First, the supervision of factual outcomes is leveraged to learn the partial representation of latent confounders corresponding to the factual outcome, i.e., $\hat{\bm{z}^y}$.
This partial representation contains information that is useful in the inference of factual outcomes. 
Specifically, we aim to learn a function $f^y:\mathds{R}^D \rightarrow \mathds{R}$ that maps the partial representation to the factual outcome.
We implement the function $f^y$ with a neural network with fully connected layers and ELU activation units (except the last layer).
Therefore, we introduce a penalty term which minimizes the mean squared error on the inferred factual outcomes as:
\begin{equation}
    \mathcal{L}^y = \frac{1}{N}\sum_{i=1}^N(f^y(\hat{\bm{z}_i^y})-y_i)^2.
    \label{eq:ly}
\end{equation}

\noindent\textbf{Observed Treatment Prediction Loss.}
Then, we utilize the observed treatment as the label to supervise the learning process of the latent confounders' treatment partial representation, i.e., $\hat{\bm{z}^t}$.
Here, an observed treatment prediction function $f^t:\mathds{R}^D\rightarrow (0,1)$ that maps the partial representation to the estimated propensity score $\hat{P}(t=1|\hat{\bm{z}}_i^t)$ is parameterized by a fully connected layer with a sigmoid activation as
\begin{equation}
    \hat{P}(t=1|\hat{\bm{z}^t}) = f^t(\hat{\bm{z}^t}) = \sigma(\mathbf{v}^T\hat{\bm{z}}_i^t+c),
    \label{eq:lt}
\end{equation}
where $\sigma$ is the sigmoid function, $\mathbf{v}$ and $c$ are the weight vector and the bias.
Then we train the partial representation $\hat{\bm{z}^t}$ by minimizing the cross-entropy loss on predicting the observed treatment:
\begin{equation}
\begin{split}
        & \mathcal{L}^{t} = -\frac{1}{N}\sum_i t_i\log(\hat{P}(t=1|\hat{\bm{z}^t})) \\ & + (1-t_i)\log(\hat{P}(t=0|\hat{\bm{z}^t})).
\end{split}
\end{equation}

\noindent\textbf{Maximizing the Mutual Information between Partial Representations.}
Intuitively, both partial representations are learned to approximate part of the information contained in the latent confounders.
Therefore, we propose to let the two partial representations agree with each other by maximizing the mutual information between the distributions of the two partial representations of latent confounders, i.e., $P(\hat{\bm{z}^t})$ and $P(\hat{\bm{z}^y})$.
Mutual information is a measure of dependence between two random variables which gauges how much the uncertainty in one variable can be reduced by knowing the value of the other one.
We know that mutual information is equivalent to the Kullback-Leibler (KL) divergence between the joint distribution and the product of the marginals~\cite{belghazi2018mutual}.
It is often quite difficult to compute mutual information with multi-dimensional continuous random variables.
Here, the Donsker-Varadhan representation of KL divergence~\cite{donsker1983asymptotic} is adopted to compute a tight lower bound of the mutual information between the distribution of the treatment partial latent confounders $P(\hat{\bm{z}^t})$ and that of the the outcome partial latent confounders $P(\hat{\bm{z}^y})$ as:
\begin{equation}
\begin{split}
        & MI(\hat{\bm{z}^t},\hat{\bm{z}^y}) = D_{KL}(P(\hat{\bm{z}^t},\hat{\bm{z}^y})||P(\hat{\bm{z}^y})\otimes P(\hat{\bm{z}^y})) = \\ & \sup_{h\in\mathcal{H}} \mathds{E}_{P(\hat{\bm{z}^t},\hat{\bm{z}^y})}[h(\hat{\bm{z}^t},\hat{\bm{z}^y})] - \log(\mathds{E}_{P(\hat{\bm{z}^y})\otimes P(\hat{\bm{z}^y})}[\exp^{h(\hat{\bm{z}^t},\hat{\bm{z}^y})}]),
\end{split}
\end{equation}
where $h:\mathds{R}^D\times\mathds{R}^D\rightarrow\mathds{R}$ is a function that maps the two partial representations to a real number and $\mathcal{H}$ denotes the space of such functions.
We can confirm that given the function $h$, the lower bound of the mutual information can be efficiently computed by sampling $(\hat{\bm{z}^t},\hat{\bm{z}^y})$ from the empirical joint distribution $P(\hat{\bm{z}^t},\hat{\bm{z}^y})$ and sampling $\hat{\bm{z}^t}$ and $\hat{\bm{z}^y}$ separately from the empirical marginals $P(\hat{\bm{z}^t})$ and $P(\hat{\bm{z}^y})$.
Here, we parameterize the function $h$ with a neural network with fully connected layers and ELU activation (except the last layer).
Then we can formulate the penalty term that maximizes the lower bound of the mutual information between the two partial representations as:
\begin{equation}
\begin{split}
    \mathcal{L}^{MI} = &- \mathds{E}_{P(\hat{\bm{z}^t},\hat{\bm{z}^y})}[h(\hat{\bm{z}^t},\hat{\bm{z}^y})] \\ &+ \log(\mathds{E}_{P(\hat{\bm{z}^y})\otimes P(\hat{\bm{z}^y})}[\exp^{h(\hat{\bm{z}^t},\hat{\bm{z}^y})}]).
\end{split}
\label{eq:lmi}
\end{equation}

\noindent Finally, we can formulate the training objective as
\begin{equation}
\underset{\bm{\theta}_{-h},\bm{\theta}_h}{\arg\min}\;\mathcal{L} = \mathcal{L}^y+\gamma\mathcal{L}^t+\zeta\mathcal{L}^{MI},
\label{eq:loss}
\end{equation}
where $\bm{\theta}_{-h}$ denotes the parameters of those components implementing the functions $g^t$, $g^y$, $f^t$ and $f^y$; while $\bm{\theta}_{h}$ signifies the parameters of the component implementing the function $h$.
The hyperparameters $\gamma$ and $\zeta$ are non-negative scalars that control the trade-off between the three penalty terms.

\subsection{Optimization}
\begin{algorithm}[t!]
	\small
	\caption{Learning the partial representations of latent confounders}
	\label{alg:1}
	\begin{algorithmic}[1]
		\renewcommand{\algorithmicrequire}{\textbf{Input:}}
		\renewcommand{\algorithmicensure}{\textbf{Output:}}
		\REQUIRE networked observational data (training set) $\left\{(\bm{x}_i,t_i,y_i)_{i=1}^N,\bm{A}\right\}$; learning rate $\eta$; hyperparameters $\gamma$ and $\zeta$; maximum number of iterations $E$; the functions $g^t$, $g^y$, $f^t$, $f^y$ and $h$.
		\ENSURE Partial representations: $\hat{\bm{z}}_i^t$ and $\hat{\bm{z}}_i^y$.
		\\ \text{Init} : Let iteration counter $e=0$; Initialize model parameters $\bm{\theta}_{-h}$ and $\bm{\theta}_h$ with Xavier initialization.
		\WHILE {$e \le E$}
		\STATE compute the partial representations $\hat{\bm{z}}_i^t$ and $\hat{\bm{z}}_i^y$ with Eq.~\eqref{eq:pr}.
		\STATE compute $\mathcal{L}^y$ and $\mathcal{L}^t$ with Eq.~\eqref{eq:ly} and Eq.~\eqref{eq:lt}.
		\STATE use $(\hat{\bm{z}}_i^t, \hat{\bm{z}}_i^y)_{i=1}^N$ as samples of the joint distribution $P(\hat{\bm{z}}^t, \hat{\bm{z}}^y)$;
		\STATE use $(\hat{\bm{z}}_i^t)_{i=1}^N,(\hat{\bm{z}}_{j(i)}^y)_{i=1}^N$ as the samples from the two marginals $P(\hat{\bm{z}}^t)$ and $P(\hat{\bm{z}}^y)$, where $j(i)$ is the $i$-th element of the permuted index vector $permute([1,...,N])$.
		\STATE compute $\mathcal{L}^{MI}$ and $\mathcal{L}$ with Eq.~\eqref{eq:lmi} and ~\eqref{eq:loss}.
		\STATE update $\bm{\theta}_{-h} \leftarrow Adam(\mathcal{L},\bm{\theta}_{-h})$.
		\STATE update $\bm{\theta}_{h} \leftarrow Adam(\mathcal{L},\bm{\theta}_{h})$.
		\ENDWHILE
	\end{algorithmic}
\end{algorithm}

Here, we describe the optimization algorithm with which the proposed framework learns the partial representations.
Algorithm~\ref{alg:1} exhibits an overview of this optimization algorithm.
In each epoch, the framework first computes the loss function $\mathcal{L}$ (Step 2-6 in Algorithm~\ref{alg:1}) in the feed-forward direction.
Then, in Step 7 and 8 of Algorithm~\ref{alg:1}, the two sets of parameters $\bm{\theta}_{-h}$ and $\bm{\theta}_h$ are updated by applying one step of gradient descent where the gradients are computed by the Adam optimizer~\cite{kingma2014adam}.

\subsection{Counterfactual Evaluation}
With the functions trained, we can compute the partial representations of any instance in the networked observational data. 
However, the counterfactual outcomes and the propensity scores inferred by the functions $f^y$ and $f^t$ can be suboptimal because each of them only uses one of the partial representations.
To overcome this issue, we propose to combine the partial representations to estimate the utility of a treatment assignment function.
In particular, we take the concatenation of them to form the representation of latent confounders as $\hat{\bm{z}}_i = concat([\hat{\bm{z}}_i^y,\hat{\bm{z}}_i^t])$.
Then the representation of latent confounders is used to train a doubly robust estimator.
%
%
First, to infer counterfactual outcomes, we follow the direct naive method in~\cite{bennett2019policy}.
In particular, for each treatment group, we simply train a neural network with fully connected layers and ELU activation (expect the last layer) with $(\hat{\bm{z}}_i)_{i=1}^N$ as the input and the factual outcomes $(y_i)_{i=1}^N$ as the label.
Second, to estimate the propensity scores, a logistic regression model is trained with $(\hat{\bm{z}}_i)_{i=1}^N$ as input and the observed treatments $(t_i)_{i=1}^N$ as the label.
Then with the two trained models and the latent confounder representations $\hat{\bm{z}}_i$, we adapt the original doubly robust estimator (Eq.~\eqref{eq:dr}) to infer the utility of a treatment assignment function $\pi$ with networked observational data as:
\begin{equation}
\begin{split}
        &\hat{\tau}(\pi) = \frac{1}{N}\sum_{i=1}^N[ \sum_t \pi^t(\bm{x}_i,\bm{A})\hat{y}_i(\hat{\bm{z}}_i,t) \\& + \hat{w}_{SNIPS}(\hat{\bm{z}}_i,t_i)(y_i-\hat{y}_i(\hat{\bm{z}}_i,t_i))],
\end{split}
\end{equation}
where $\hat{y}_i(\hat{\bm{z}}_i,t_i)$ is the outcome inferred by the simple direct method. 
In terms of the sample weights, we adopt the self-normalized inverse propensity scoring ($\hat{w}_{SNIPS}$) to avoid the extreme values and reduce variance~\cite{swaminathan2015self}.
The self-normalized inverse propensity scoring weights are computed as:
\begin{equation}
\begin{split}
       \hat{w}_{SNIPS}(\hat{\bm{z}}_i,t_i) & = \frac{\hat{w}_{IPS}(\hat{\bm{z}}_i,t_i)}{\sum_{i=1}^N\hat{w}_{IPS}(\hat{\bm{z}}_i,t_i)},
\end{split}
\label{eq:new_dr}
\end{equation}
where $\hat{w}_{IPS}(\hat{\bm{z}}_i,t_i) = \frac{\pi^{t_i}(\bm{x}_i,\bm{A})}{\hat{P}(t=t_i|\hat{\bm{z}}_i)}$. The probability for instance $i$ to receive the treatment $t_i$, $\hat{P}(t=t_i|\hat{\bm{z}}_i)$, is inferred by the logistic regression model.
%

%
\section{Experiments}

In this section, we investigate whether network information among observational data can help improve counterfactual evaluation through extensive experiments.
%
%
%
%

\subsection{Dataset Description}
In real-world situations, only the factual outcome of each instance is observable.
For example, we can observe the potential outcome $y_i^1$ of the $i$-th instance iff $t_i=1$.
As a result, it is extremely challenging to collect data with ground truth of counterfactual outcomes. 
Therefore, we follow~\cite{veitch2019using,guo2019learning} to synthesize the treatments and outcomes based on the observed features and network information which are extracted from two real-world datasets.
%
%
Specifically, we introduce two networked observational datasets for evaluating the utility of treatment assignment functions. 
Based on the observed features and the network structures, we introduce the data generating process which synthesizes treatments and outcomes.
To reflect real-world situations, we consider hidden confounders and unknown edge weights in the process.
We fully cover the steps to reproduce the semi-synthetic datasets from the publicly available social network datasets, BlogCatalog and Flickr.

\noindent\textbf{BlogCatalog} (BC) is a social media website where users post blogs.
In this dataset, each instance is a blogger. Each edge presents the friendship between two bloggers.
The original features are the bag-of-words representation of the keywords of the articles posted by a blogger.
Here, the task is to learn a treatment assignment function which automatically determines to promote a blogger's article more on mobile devices or desktops such that users' opinion is optimized.
A treatment assignment function decides to promote a blogger's article more on mobile devices or desktops.
We extend the original BlogCatalog dataset~\cite{li2016robust} by synthesizing (a) the outcomes -- readers' opinions on bloggers; and (b) the treatments -- readers' device preference.
Similar to the News dataset~\cite{johansson2016learning,schwab2018perfect} that are widely used in causal inference literature, the following assumptions are made: (1) Readers either read on mobile devices or desktops. We say a blogger get treated (controlled) if her blogs are more popular on mobile devices (desktops). 
(2) Readers prefer to read certain topics from mobile devices, others from desktops.
(3) The latent confounders of a blogger are determined by a function of her and her neighbors' topics.
(4) The latent confounders of a blogger causally influence both readers' preference of reading devices (treatment) and readers' opinion (outcome).
%
%
Based on these assumptions, we train a topic model on a large set of documents to synthesize treatments and outcomes.
Then we define the centroid of each treatment group in the topic space: (i) we randomly sample a blogger whose topic distribution becomes the centroid of the treated group, denoted by $\bar{r}^1$; (ii) we let the centroid of the controlled, $\bar{r}^0$, be the average topic distribution of all bloggers.
Then we introduce how the treatments and outcomes are synthesized based on the similarity between the topic distributions of bloggers and the two centroids.
Let $r(\bm{x}_i)$ be the topic distribution of the $i$-th blogger's description, we model the readers' preference of browsing devices on the blogger's content: 
\begin{equation}
P(t=1|\bm{x}_i,\tilde{\bm{A}}) = \frac{\exp(p_i^1)}{\exp(p_i^1)+\exp(p_i^0)},
\end{equation}
%
where $p_i^t$ is calculated as:
\begin{equation}
    \begin{split}
        & p_t^i =  \kappa_1 r(\bm{x}_i)^T \bar{r}^t+\kappa_2(\tilde{\bm{A}}r(\bm{x}_j))^T\bar{r}^t.
    \end{split}
\end{equation}


%
\noindent where $t\in\{0,1\}$.
For blogger $i$, the first term on RHS represents the confounding bias caused by the topics of herself. The second term on RHS signifies that caused by the topics of her neighbors. $\kappa_1\ge 0$ and $\kappa_2 \ge 0$ control the strength of these two terms.
%
When $\kappa_1=\kappa_2=0$ the treatment assignment is random and the greater the value $\kappa_1$ and $\kappa_2$ are, the less the treatment assignment is, and therefore, the more significant the confounding bias is.
We let $\tilde{\bm{A}}$ denote the normalized weighted adjacency matrix, where each entry $\tilde{\bm{A}}_{ij}$ denotes the importance of an edge with related to the influence on confounding bias.
To reflect the fact that in social networks the edge weights are unknown, only the unweighted adjacency matrix $\bm{A}$ is observable in the data.  However, the unobserved weighted adjacency matrix $\tilde{\bm{A}}$ is the one that influences the values of treatments and outcomes.  
%
Thus, an ideal causal inference approach needs to catch the weights of each edge.
%
If $\bm{A}_{ij}=1$, then we sample $\tilde{\bm{A}}_{ij} = \tilde{\bm{A}}_{ji}\sim \text{Uniform}(0.1,1)$; otherwise, we set $\tilde{\bm{A}}_{ij} = \tilde{\bm{A}}_{ji} = 0$.
%
%
%
%
Then the raw outcomes of the $i$-th blogger are simulated as:
\begin{equation}
y_i^{raw}(t)=(1-t)p_0^i+ tp_1^i+\epsilon.
\end{equation}
%
The noise $\epsilon$ is sampled from a zero-mean Gaussian distribution with standard deviation $0.01$.
Then we normalize the raw outcomes $y_i(t)$ with:
\begin{equation}
	y_i(t) = \frac{y_i^{raw}(t) - \mu(\bm{y}^{raw})}{\sigma'(\bm{y}^{raw})},
\end{equation} 
where $\mu(\bm{y}^{raw})$ and $\sigma'(\bm{y}^{raw})$ signify the mean and standard deviation of all raw outcomes. 
%
%
In this work, we set $\kappa_1=1$ and vary $\kappa_2 \in \{1,2\}$.
Meanwhile, 50 LDA topics are learned from the training corpus. 
Then we reduce the vocabulary by taking union of the most probable 100 words from each topic, which results in 2,173 bag-of-word features.
%
%
%


\noindent\textbf{Flickr} is an online community where users share images and videos. 
Each instance is a user and each edge is the friendship between two users. 
The original features of each user are the tags of interest.
We adopt the same settings and assumptions as we do for the BC datasets.
Thus, we study the ITE of being viewed on mobile devices on readers' opinions on the user.
We learn 50 topics from the training corpus using LDA and concatenate the top 25 words of each topic which reduces the feature dimension to 1,210.
Meanwhile, we set the parameters the same as the BC dataset.

\begin{table*}
	\caption{Statistics of the datasets}
	\scriptsize
	\label{tab:datasets}
	\centering
	
	\begin{tabular}{|c|c|c|c|c|c|c|}
		\hline
		Dataset & Instances & Edges & Features  & $\kappa_2$ & Treated Instances & Instances with $y_i^1>y_i^0$ \\
		\hline
		\multirow{2}{*}{BC} & \multirow{2}{*}{5,196} & \multirow{2}{*}{173,468} & \multirow{2}{*}{8,189} & 1 & 2579.5 $\pm$ 29.891 & 1030.1 $\pm$ 331.31\\
		&&&& 2 & 2448.6 $\pm$ 539.687 & 2031.1 $\pm$ 1149.696  \\ \hline
		\multirow{2}{*}{Flickr} & \multirow{2}{*}{7,575} & \multirow{2}{*}{239,738} & \multirow{2}{*}{12,047} & 1 & 3700.8 $\pm$ 156.873 & 2708.3 $\pm$ 745.03
 \\
		&&&& 2 & 3859.4 $\pm$ 218.072 & 3182.1 $\pm$ 588.958\\ \hline
	\end{tabular}
\end{table*}

Table~\ref{tab:datasets} presents the statistics of the two semi-synthetic datasets under various settings. 
The average and standard deviation of the number of treated instances and the number of instances that satisfy $y_i^1>y_i^0$ are calculated over the 10 simulations under each setting of parameters.
They vary because the true edge weights are randomly sampled from the uniform distribution $\text{Uniform}(0.1,1)$.

\subsection{Experimental Settings.}
We randomly split each dataset into training (60\%), validation (20\%) and test sets (20\%) and report the results of the test sets for 10 runs on each simulation.
%
%
%
Grid search is applied to find optimal hyperparameters, details can be found in Appendix.
For evaluation, we adapt those in~\cite{bennett2019policy,zou2019focused} to a class of the treatment assignment functions which take both observed features and network information as input.
%
%
The treatment assignment functions with random weights are considered:
\begin{equation}
	\pi_{rw}^t(\bm{x}_i,\bm{A}) = \frac{\exp({\bm{\psi}^t}^T\bm{x}_i+\frac{1}{|\mathcal{N}(i)|}\sum_{j\in\mathcal{N}(i)}{\bm{\delta}^t}^T\bm{x}_j)}{\sum_{t}\exp({\bm{\psi}^t}^T\bm{x}_i+\frac{1}{|\mathcal{N}(i)|}\sum_{j\in\mathcal{N}(i)}{\bm{\delta}^t}^T\bm{x}_j)},
\end{equation}
where $\mathcal{N}(i)$ is the set of neighbors of instance $i$ in the network $\bm{A}$. The random weights are obtained as $\bm{\psi}^1,\bm{\delta}^1\sim 2Bern(\bm{0.5})-\bm{1}$ and $\bm{\psi}^0 = -\bm{\psi}^1,\bm{\delta}^0 = -\bm{\delta}^1$.
For these treatment assignment functions, the ground truth utilities are obtained with Eq.~\eqref{eq:utility}.

To corroborate the effectiveness of CONE, it is evaluated against the following state-of-the-art kernel based methods, neural network based methods, and classic methods:
\begin{itemize}
	%
	
	\item Optimal Kernel Balancing (OKB)~\cite{bennett2019policy} is the state-of-the-art kernel based weighted estimator which minimizes an adversarial balance objective.
	
	\item Inverse propensity scoring (IPS-X)~\cite{bottou2013counterfactual} is a weighted estimator which fits a propensity scoring model using observed features. Specifically, a logistic regression model is trained with supervision of the observed treatments.
	
	\item Self-normalized inverse propensity scoring (SNIPS-X)~\cite{swaminathan2015self} is a variant of the weighted estimator IPS-X where self-normalized weights are employed.
	
	\item The direct method~\cite{qian2011performance} estimates the utility of a treatment assignment function through the inference of counterfactual outcomes (Eq~\eqref{eq:dm}). We consider three models that can infer counterfactual outcomes: OLS1, OLS2~\cite{louizos2017causal}, and the simple direct method using a neural network model (DM-X)~\cite{bennett2019policy}.
	
	\item A doubly robust estimator~\cite{dudik2011doubly} combines a direct method and the inverse propensity scoring (Eq.~\eqref{eq:dr}).
	Here, we consider the combination of each aforementioned direct method (OLS1, OLS2 or DM-X) and the IPS-X method. We refer to them as DR-OLS1, DR-OLS2, and DR-DM-X.
	
\end{itemize}
%
As this is the first work utilizing network information for counterfactual evaluation, there is no baseline that naturally incorporates network information.
We also tried to concatenate the adjacency matrix to the original features to allow baselines utilize the network information for a fair comparison.
However, such an approach cannot improve the performance of baselines due to the high dimensionality and sparsity of the network information.

Then, we formally present the two evaluation metrics, root mean squared error (RMSE) and mean absolute error (MAE) as
$RMSE = \sqrt{\frac{1}{K}\sum_{k=1}^K(\hat{\tau}_k(\pi)-\tau_k(\pi))^2}$ and $MAE = \frac{1}{K}\sum_{k=1}^K |\hat{\tau}_k(\pi)-\tau_k(\pi)|,
$
where $K$ is the number of simulations.

\subsection{Effectiveness}
Here, experimental results are shown to corroborate the effectiveness of the proposed framework.
Table~\ref{tab:res} shows the empirical results evaluated on the BC and Flickr datasets with $\kappa_1 = 1$ and $\kappa_2 \in\{1,2\}$.
The following observations are made from these experimental results:
\begin{itemize}
    \item The proposed framework CONE results in better performance than the state-of-the-art baseline methods consistently on both datasets under various settings.
    One-tailed T-tests show that the results of CONE are significantly better with a significance level of 0.05.
    
    \item Measured by the increase in both error metrics, the performance of CONE worsens less than other methods when the hidden confounding effect grows (from $\kappa_2=1$ to $\kappa_2=2$). This further verifies that capturing the patterns of hidden confounders from the network structure with the combined partial representations helps counterfactual evaluation of treatment assignment functions.
\end{itemize}

\begin{table*}[]
\scriptsize
\caption{Experimental results corroborating the effectiveness of CONE}
\label{tab:res}
\centering
\begin{tabular}{|l|l|l|l|l|l|l|l|l|}
\hline
 & \multicolumn{4}{l|}{BlogCatalog} & \multicolumn{4}{l|}{Flickr} \\ \hline
 & \multicolumn{2}{l|}{$\kappa_2=1$} & \multicolumn{2}{l|}{$\kappa_2=2$} & \multicolumn{2}{l|}{$\kappa_2=1$} & \multicolumn{2}{l|}{$\kappa_2=2$} \\ \hline
 & RMSE & MAE & RMSE & MAE & RMSE & MAE & RMSE & MAE \\ \hline
\textbf{CONE (ours)} & \textbf{0.034} & \textbf{0.026} & \textbf{0.037} & \textbf{0.027} & \textbf{0.014} & \textbf{0.011} & \textbf{0.014} & \textbf{0.012} \\ \hline
OKB & 0.141 & 0.135 & 0.150 & 0.143 & 0.073 & 0.063 & 0.093 & 0.083 \\ \hline
IPS-X & 0.042 & 0.039 & 0.089 & 0.074 & 0.018 & 0.016 & 0.030 & 0.027 \\ \hline
SNIPS-X & 0.042 & 0.038 & 0.089 & 0.074 & 0.018 & 0.017 & 0.029 & 0.027 \\ \hline
DM-X & 0.229 & 0.229 & 0.241 & 0.239 & 0.099 & 0.097 & 0.117 & 0.114 \\ \hline
OLS1 & 0.302 & 0.301 & 0.347 & 0.346 & 0.144 & 0.143 & 0.168 & 0.167 \\ \hline
OLS2 & 0.275 & 0.274 & 0.308 & 0.304 & 0.139 & 0.139 & 0.162 & 0.161 \\ \hline
DR-DM-X & 0.041 & 0.034 & 0.071 & 0.060 & 0.019 & 0.018 & 0.028 & 0.026 \\ \hline
DR-OLS1 & 0.042 & 0.039 & 0.089 & 0.074 & 0.018 & 0.016 & 0.030 & 0.027 \\ \hline
DR-OLS2 & 0.047 & 0.041 & 0.090 & 0.078 & 0.019 & 0.017 & 0.031 & 0.028 \\ \hline
\end{tabular}
\end{table*}


\subsection{Parameter Study}
\begin{figure}[tbh!]
 \centering
 \begin{minipage}{0.24\textwidth}
 \centering
  \subfigure[RMSE of BC ($\kappa_2=2$)]
 {\includegraphics[width=\textwidth]{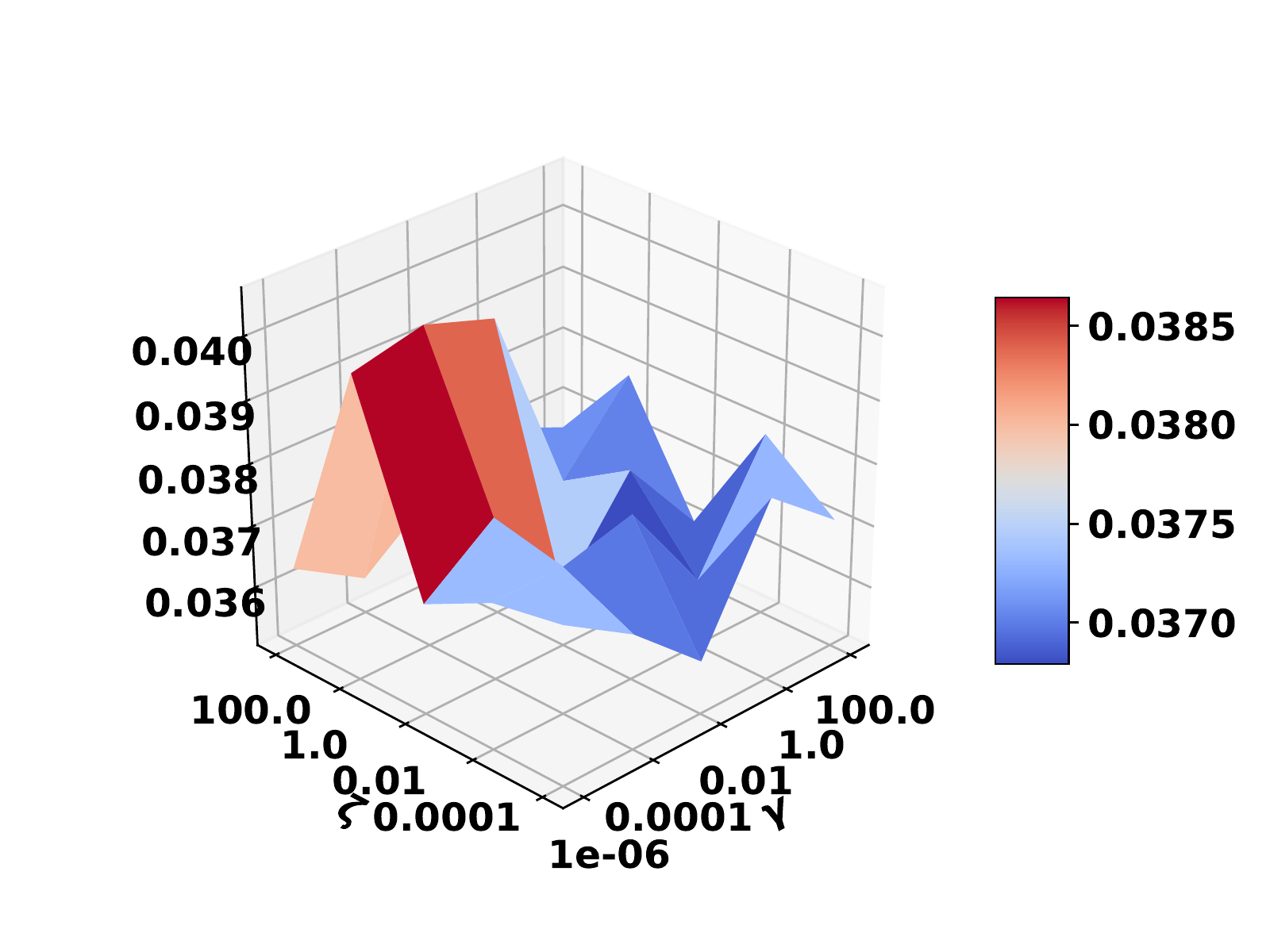}}
 \end{minipage}
  \hfil
  \begin{minipage}{0.24\textwidth}
  \centering
  \subfigure[MAE of BC ($\kappa_2=2$)]
  {\label{fig:proposed}\includegraphics[width=\textwidth]{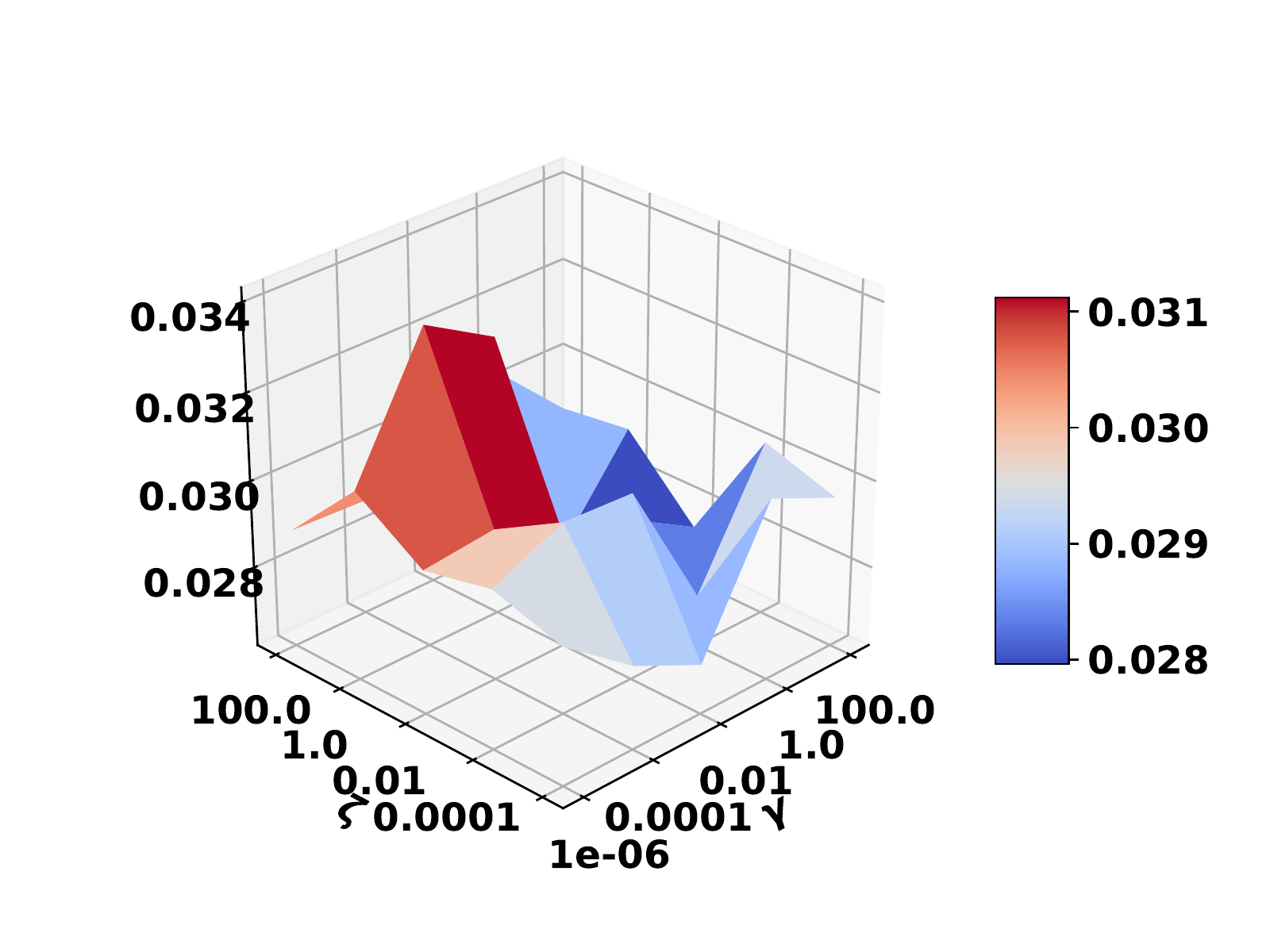}}
  \end{minipage}
\hfil
 \begin{minipage}{0.24\textwidth}
	\centering
	\subfigure[RMSE of Flickr ($\kappa_2=2$)]
	{\includegraphics[width=\textwidth]{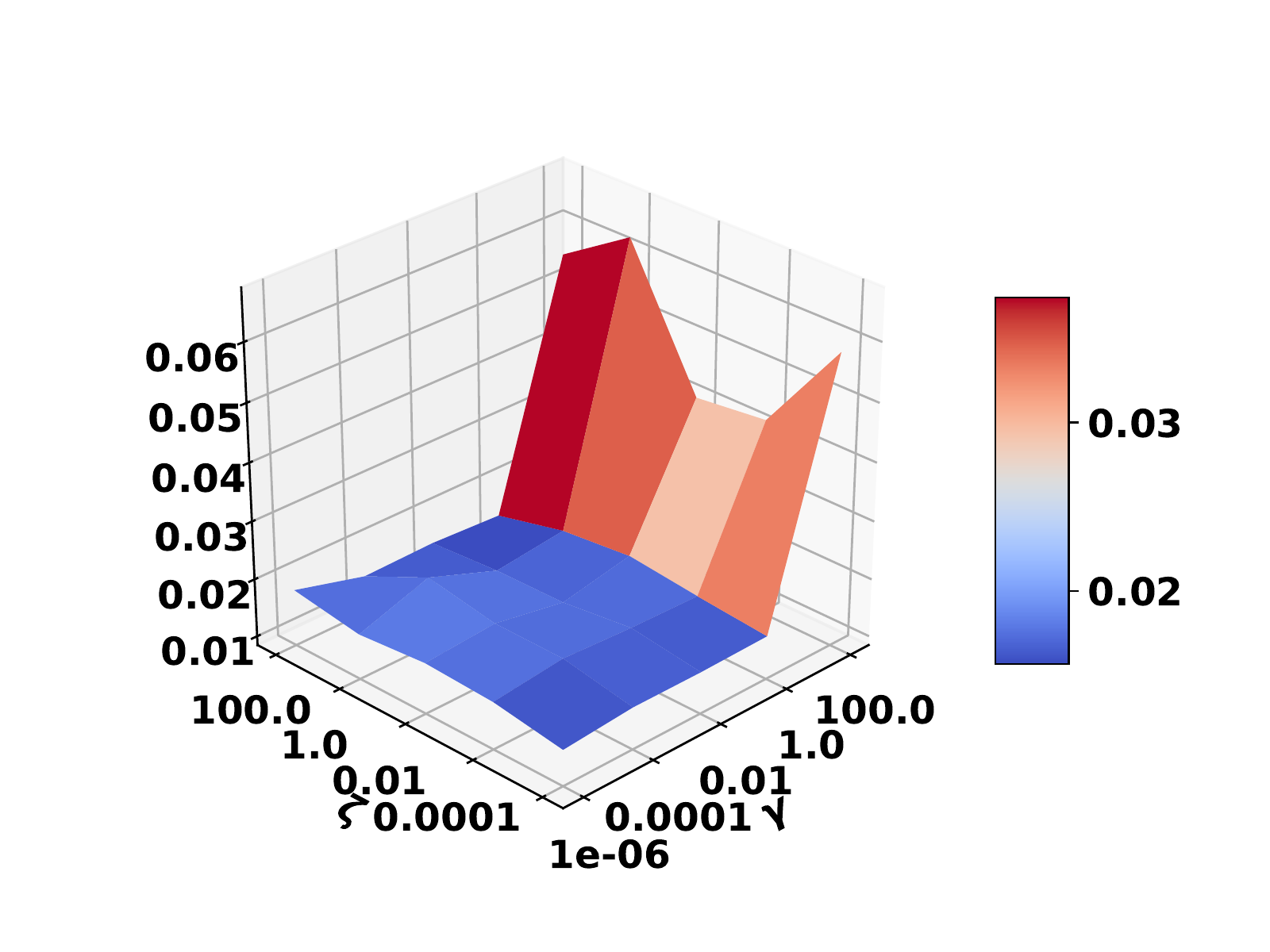}}
\end{minipage}
\hfil
\begin{minipage}{0.24\textwidth}
	\centering
	\subfigure[MAE of Flickr ($\kappa_2=2$)]
	{\label{fig:proposed}\includegraphics[width=\textwidth]{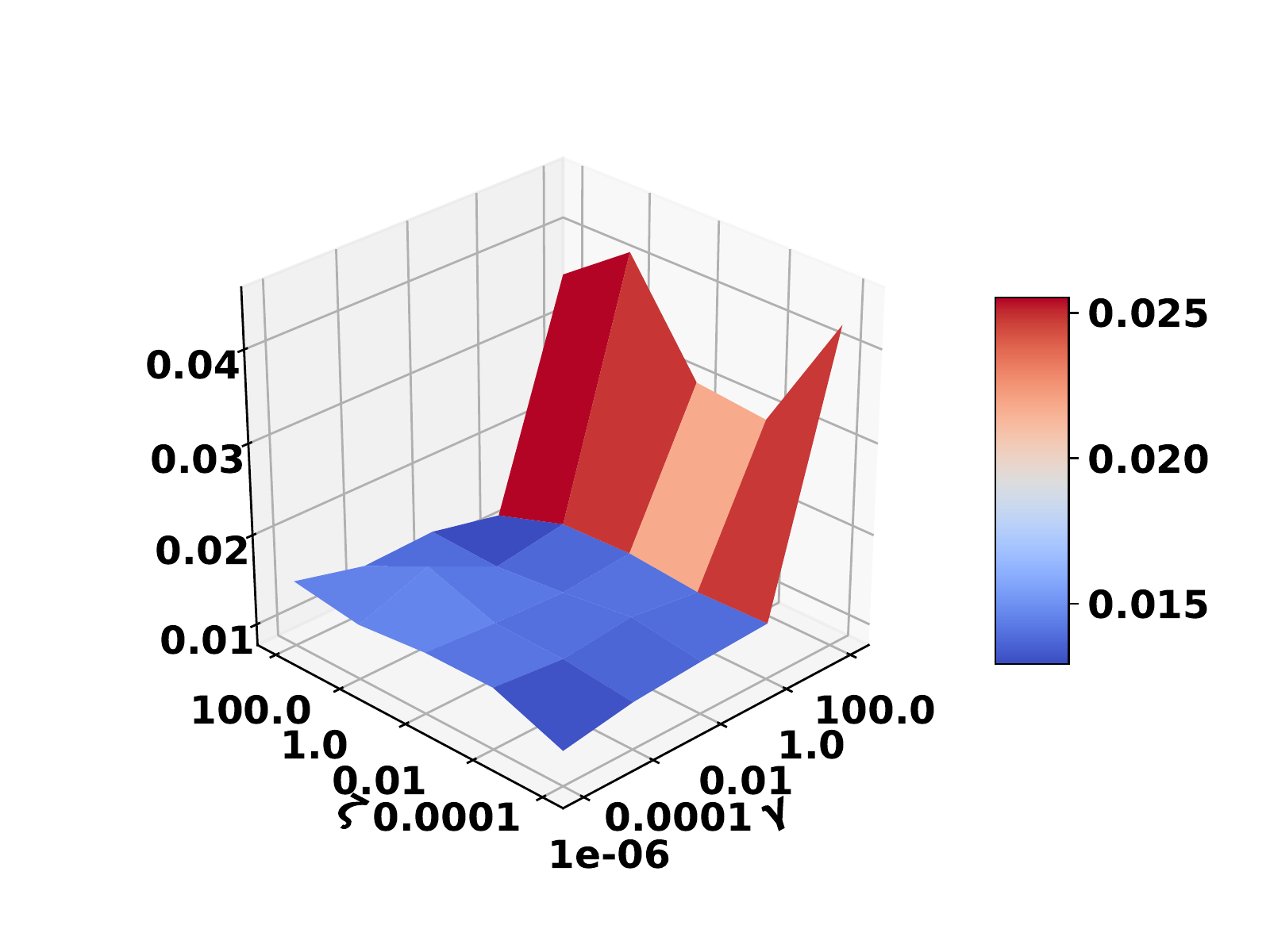}}
\end{minipage}

 \caption{Parameter study results}

 \label{fig:param_study}
 \end{figure}

Here, we investigate the influence of the hyperparameters $\gamma$ and $\zeta$ on the performance of CONE.
Note that $\gamma$ controls the penalty on the predictions of observed treatments based on the related partial representation.
And $\zeta$ determines to what extent the two partial representations agree with each other. 
The detailed setup of the parameter study can be found in the Appendix.
We set $\gamma$ and $\zeta$ in the range of $\{10^{-6},10^{-4},10^{-2},1,100\}$.
Fig.~\ref{fig:param_study} presents the experimental results on the BC and Flickr datasets with $\kappa_1=1$ and $\kappa_2=2$.
We can observe that when $\gamma\in[1,100]$ and $\zeta\in[0.01,1]$, CONE achieves the best performance on the BC dataset ($\kappa_2=2$).
For the Flickr dataset ($\kappa_2=2$), CONE consistently performs well when $\gamma\in[10^{-6},1]$ and $\zeta\in[10^{-6},100]$.
Similar results can be obtained with other settings.
We can conclude that CONE maintains stable performance by varying the hyperparameters in a wide range, which is often desired in real-world applications.

\section{Related Work}
Counterfactual evaluation methods for i.i.d. data consist of three major categories: direct methods~\cite{qian2011performance}, weighted estimators~\cite{kallus2018balanced,bennett2019policy,swaminathan2015counterfactual,swaminathan2015self}, and doubly robust estimators~\cite{dudik2011doubly,athey2017efficient}.
Directed methods achieve counterfactual evaluation by inferring counterfactual outcomes.
However, existing direct methods are known to suffer from biased estimates~\cite{beygelzimer2008importance}.
This is mainly because that they rely on the unconfoundedness assumption. Moreover, the supervision of the observed treatments remains to be utilized.
Weighted estimators avoid the problem of inferring counterfactual outcomes.
Instead, they estimate the utility of treatment assignment functions through a weighted average of observed outcomes.
In particular, a sample weight is learned for each instance.
The goal is to let the reweighted factual outcomes approximate their counterparts that would have been observed if the treatments had been assigned by the function to be evaluated.
Inverse propensity scoring (IPS)~\cite{kitagawa2018should,hirano2003efficient} is the most widely adopted strategy for reweighting.
IPS estimators can suffer from the issue of high variance when the estimated propensity scores take extreme values.
Therefore, a series of clipping and normalization based methods~\cite{bottou2013counterfactual,swaminathan2015counterfactual,swaminathan2015self} have been proposed to mitigate this issue.
However, IPS estimators' performance is still limited by the accuracy of estimated propensity scores.
To combine the advantages of the two types of methods, doubly robust estimators are proposed~\cite{chernozhukov2018double,bang2005doubly}.
Each doubly robust estimator consists of a direct method and a weighted estimator.
Previous work~\cite{dudik2011doubly,chernozhukov2018double} has shown doubly robust estimators can maintain good performance even if either its direct method or its IPS estimator suffers from large bias.
Different from the aforementioned work, this work investigates the effectiveness of incorporating network information in counterfactual evaluation.

\section{Conclusion}
In this work, we study the problem of counterfactual evaluation in networked observational data.
In particular, we investigate the hypothesis that utilizing network information will help handle hidden confounders in counterfactual evaluation.
We propose a novel framework, CONE, which leverages the network information along with the observed features to mitigate hidden confounding effects for counterfactual evaluation.
Empirical results from extensive experiments show the effectiveness of CONE and verify that incorporating network information indeed helps us control hidden confounders in the task of counterfactual evaluation. 

\bibliographystyle{siamplain}
\bibliography{sdmbib}

\end{document}